\documentclass[runningheads]{llncs}
\usepackage{times}  % DO NOT CHANGE THIS
\usepackage{helvet} % DO NOT CHANGE THIS
\usepackage{courier}  % DO NOT CHANGE THIS
\usepackage[hyphens]{url}  % DO NOT CHANGE THIS
\usepackage{graphicx} % DO NOT CHANGE THIS
\usepackage{graphicx}  % DO NOT CHANGE THIS
\usepackage{times}
\usepackage{graphicx}     
\usepackage{amsmath}
\usepackage{txfonts}
\usepackage[ruled,vlined]{algorithm2e}

\DeclareMathAlphabet{\mathitbf}{OML}{cmm}{b}{it}

\newcommand{\sub}{_}
\def\su{^}
\newcommand{\real}{{\mathbb{R}}}

\newcommand{\gt}{>}

\newcommand{\B}{{\cal B}}
\renewcommand{\H}{{\cal H}}

\newcommand{\D}{{\cal D}}

\renewcommand{\L}{{\cal L}}

\newcommand{\M}{{\cal M}}
\newcommand{\N}{{\cal N}}

\newcommand{\set}[1]{\left\{ #1 \right\}}

\begin{document}
%
% \title{Symbolic Logic meets Machine Learning: \\ The Case of Infinite Domains\thanks{
% %This is a report accompanying the tutorial given at \textit{The 14th International Conference on Scalable Uncertainty Management}, 2020.
% The author was supported by a Royal Society University Research Fellowship.}}

\title{Symbolic Logic meets Machine Learning: \\ A Brief Survey in Infinite Domains\thanks{
The author was supported by a Royal Society University Research Fellowship. He is  grateful to Ionela G. Mocanu, Paulius Dilkas and Kwabena Nuamah for their feedback.}}

%
%\titlerunning{Abbreviated paper title}
% If the paper title is too long for the running head, you can set
% an abbreviated paper title here
%
\author{
Vaishak Belle\inst{}
}

\authorrunning{V. Belle}
\institute{University of Edinburgh \&  Alan Turing Institute, UK \\ 
\email{vaishak@ed.ac.uk}} 

\maketitle              % typeset the header of the contribution
\begin{abstract}

The tension between deduction and induction is perhaps the most fundamental issue in areas such as philosophy, cognition and artificial intelligence (AI). The deduction camp concerns itself with questions about the expressiveness of formal languages for capturing knowledge about the world, together with proof systems for reasoning from such knowledge bases. The learning camp attempts to generalize from examples about partial descriptions about the world. In AI, historically, these camps have loosely divided the development of the field, but advances in cross-over areas such as statistical relational learning, neuro-symbolic systems, and high-level control have illustrated that the dichotomy is not very constructive, and perhaps even ill-formed. 

In this article, we survey work that provides further evidence for the connections between logic and learning. Our narrative is structured in terms of three strands:  logic versus learning, machine learning for logic, and logic for machine learning, but naturally, there is considerable overlap. We place an emphasis on the following ``sore"  point: there is a common  misconception that logic is for discrete properties, whereas probability theory and
machine learning, more generally, is for continuous properties. We report on results that challenge this view on the limitations of logic, and 
 expose the role that logic can play for learning in infinite domains.

%\keywords{Logic and learning  \and Second keyword \and Another keyword.}

\end{abstract}
\section{Introduction}

The tension between \textit{deduction} and \textit{induction} is perhaps the most fundamental issue in areas such as philosophy, cognition and artificial intelligence (AI). The deduction camp concerns itself with questions about the expressiveness of formal languages for capturing knowledge about the world, together with proof systems for reasoning from such knowledge bases. The learning camp attempts to generalize from examples about partial descriptions about the world. In AI, historically, these camps have loosely divided the development of the field, but advances in cross-over areas such as \textit{statistical relational learning} \cite{raedt2016statistical,an-introduction-to-statistical-relational-learning}, \textit{neuro-symbolic systems}  \cite{garcez2019neural,de2019neuro,lamb2020graph}, and \textit{high-level control} \cite{cognitive-robotics,integrated-task-and-motion-planning} have illustrated that the dichotomy is not very constructive, and perhaps even ill-formed. Indeed, logic emphasizes high-level reasoning, and encourages  structuring the world in terms of objects, properties, and relations. In contrast, much of the inductive machinery  assume  random variables to be independent and identically distributed, which can be problematic when attempting to exploit symmetries and causal dependencies between groups of objects. 
% So, by weakening the truth of logical statements (via probabilities or similar), one obtains closely related mathematical  \cite{russel}. 
But the threads connecting logic and learning go deeper, far beyond the apparent flexibility that logic offers for modeling relations and hierarchies  in  noisy domains. At a conceptual level, for example, although there is much debate  about what precisely commonsense knowledge might look like, it is widely acknowledged that concepts such as time, space, abstraction and causality are essential \cite{marcus2019rebooting,zellers2018swag}.  In that regard, (classical, or perhaps non-classical) logic can provide the formal machinery to reason about  such concepts in a rigorous way. At a pragmatic level, despite the success of methods such as deep learning, it is now increasingly recognized that owing to a number of reasons, including model re-use, transferability, causal understanding, relational abstraction, explainability and data efficiency, those methods need to be further augmented with logical, symbolic  and/or programmatic artifacts  \cite{bunel2018leveraging,xu2019can,evans2018learning}. Finally, for building  intelligent agents, it is recognized that low-level, data-intensive, reactive computations needs to be tightly integrated with high-level, deliberative computations \cite{cognitive-robotics,integrated-task-and-motion-planning,manhaeve2018deepproblog}, the latter possibly also engaging in hypothetical and counterfactual reasoning. 
% If we downplay the emphasis on low-level sensory modules
Here, a parallel is often drawn to Kahneman's so-called \textit{System 1}  versus \textit{System 2} processing in human cognition \cite{kahneman2011thinking}, in the sense that experiential and reactive processing (learned behavior) needs to be coupled with cogitative processing  (reasoning, deliberation and introspection) for sophisticated machine intelligence.    

The purpose of this article is not to resolve this debate, but rather provide further evidence for the connections  between logic and learning. In particular, our narrative is inspired by a recent  symposium on logic and learning \cite{benedikt2020logic}, where the landscape was structured in terms of three strands: {\it

\begin{enumerate}
\item \textbf{Logic vs. Machine Learning}, including the study of problems that can be solved using either logic-based techniques or via machine learning, $\ldots$; 
\item \textbf{Machine Learning for Logic}, including the learning of logical artifacts, such as formulas, logic programs, $\ldots$; and 
\item \textbf{Logic for Machine Learning}, including the role of logics in delineating the boundary between tractable and intractable learning problems,  $\ldots,$  and the use of logic as a declarative framework for expressing machine learning constructs.
\end{enumerate}

}{} 

In this article, we particularly focus on the following ``sore" point: there is a common misconception that logic is for discrete properties, whereas probability theory and machine learning, more generally, is for continuous properties. It is true that logical formulas are discrete structures, but they can very easily also express properties about countably infinite or even uncountably  many objects.  Consequently, in this article we survey some recent results that tackle the integration of logic and learning in infinite domains. In particular, in the context of the above three strands, we report on the following developments. On  (1), we discuss approaches for logic-based probabilistic inference in  continuous domains. On  (2), we cover approaches for learning logic programs in continuous domains, as well as learning  formulas that represent countably infinite sets of objects. Finally, on (3), we discuss attempts to use logic as a declarative framework for common tasks in machine learning over discrete and continuous features, as well as using logic as  a meta-theory to consider notions such as the \textit{abstraction} of a probabilistic model.

{We remark that this survey is undoubtedly a biased view, as the area of research is large, but we do attempt to briefly cover the major threads. Readers are encouraged to refer to discussions in \cite{benedikt2020logic,an-introduction-to-statistical-relational-learning,raedt2016statistical}, among others, to get a sense of the breadth of the area.
}  

% proceed as follows. We first discuss a recent extension to logic-based probabilistic inference for  continuous domains. Then we turn to (2), and cover approaches for learning logic programs in continuous domains, as well as learning  formulas that represent countably infinite sets of objects. Finally, we turn to (3) and

\section{Logic vs. Machine Learning} % (fold)
\label{sec:logic_vs_machine_learning}

To appreciate the role and impact of logic-based solvers for machine learning systems, it is perhaps useful to consider the core computational problem underlying (probabilistic) machine learning: the problem of inference, including evaluating the partition function  (or conditional probabilities) of a probabilistic graphical model such as a Bayesian network.

When leveraging Bayesian networks for machine learning tasks \cite{probabilistic-graphical-models---principles}, the networks are often learned using local search to maximize a likelihood or a Bayesian quantity. For example, given data \( \D \) and the current guess for the network \( \N \), we might estimate the ``goodness'' of the guess by means of a score: \( {\it score}(\N ,\D) \propto \log \Pr(\D\mid \N) - {\it size}(\N) \). That is, we want to maximize the fit of the data wrt the current guess, but we would like to penalize the model complexity, to avoid overfitting. Then, we would opt for a second guess \( \N' \) only if \( {\it score}(\N',\D) \gt {\it score}(\N,\D) \). Needless to say, even with a reasonable local search procedure, the most significant  computational effort   here is that of  probabilistic inference. 

Reasoning in  such networks becomes especially challenging with logical syntax. The prevalence of large-scale social networks, machine reading domains, and other types of relational knowledge bases has led to numerous formalisms that borrow the syntax of predicate logic for probabilistic modeling \cite{tuffy:-scaling-up-statistical-inference,markov-logic-networks,first-order-probabilistic-inference,probabilistic-databases}. This has led to a large family of solvers for the \emph{weighted model counting} (WMC) problem \cite{model-counting,on-probabilistic-inference-by-weighted-model}. The idea is this: given a Bayesian network, a relational Bayesian network,  a factor graph, or a probabilistic program  \cite{problog2:-from-probabilistic-programming}, one considers an encoding of the formalism as a \emph{weighted propositional theory}, consisting of a propositional theory \( \Delta \) and a weight function \( w \) that maps atoms in \( \Delta \) to \( \real\su +  \).  Recall that SAT is the problem of finding an assignment to such a \( \Delta, \) whereas \#SAT counts the number of assignments for \( \Delta. \) WMC extends \#SAT by  computing the sum of the weights of all assignments: that is, given a set of models \( \M(\Delta) = \set{ M \mid M \models \Delta} \), 
we evaluate the quantity \(
	W(\Delta) = \sum\sub {M \in \M(\Delta)} w(M)
\) 
where \( w(M) \)  is factorized in terms of the atoms true at \( M. \) To obtain the conditional probability of a query \( q \) against evidence \( e \) (wrt the theory \( \Delta \)), we define \( \Pr(q\mid e) = W(\Delta\land q \land e) / W(\Delta\land e). \)

% \( \set{ w(M) \mid M \models \Delta} \), where \( w(M) \)  is factorized in terms of the atoms true at \( M. \)

The popularity of WMC can be explained as follows. Its formulation elegantly decouples the logical or symbolic representation from the  numeric representation, which is encapsulated in the weight function. When building solvers, this allows us to reason about logical equivalence and reuse SAT solving technology (such as constraint propagation and clause learning). WMC also makes it more natural to reason about deterministic, hard constraints in a probabilistic context   \cite{on-probabilistic-inference-by-weighted-model}. Both exact solvers, based on knowledge compilation \cite{new-advances-in-compiling-cnf-to-decomposable-negation}, as well as approximate solvers \cite{distribution-aware-sampling-and-weighted-model} have emerged in the recent years, as have lifted techniques \cite{lifted-inference-and-learning-in-statistical} that exploit the relational syntax during inference  (but in a finite domain setting). For ideas on generating such representations randomly to assess scalability and compare inference algorithms, see \cite{dilkas2020generating}, for example. 

On the point of modelling finite vs infinite properties, 
%With regards to the finitary nature of this development, 
note that owing to the underlying propositional language, the formulation is limited to discrete random variables. A similar observation can be made for SAT, which for the longest time could only be applied in discrete domains. This changed with the increasing popularity of \emph{satisfiability modulo theories} (SMT) \cite{satisfiability-modulo-theories}, which enable us to, for example, reason about the satisfiability of linear constraints over the rationals. Extending earlier insights on piecewise-polynomial weight functions  \cite{inference-in-hybrid-bayesian-networks,symbolic-variable-elimination-for-discrete}, the formulation of \emph{weighted model integration}  (WMI)  was proposed in \cite{probabilistic-inference-in-hybrid-domains}. WMI extends WMC by leveraging the idea that SMT theories can represent mixtures of Boolean and continuous variables: for example, a formula such as \( p \land (x\gt 5) \) denotes the logical conjunction of a Boolean variable \( p \) and a real-valued variable \( x \) taking values greater than 5. For every assignment to the Boolean and continuous variables, the WMI problem defines a weight. The total WMI is computed by integrating these weights over the domain of solutions to \( \Delta \), which is a mixed discrete-continuous (or simply \textit{hybrid}) space. Consider, for example, the special case when \( \Delta \)  has no Boolean variables, and the weight of every model is 1. Then, the WMI simplifies to computing the volume of the polytope encoded in \( \Delta \). When we additionally allow for Boolean variables in \( \Delta \), this special case becomes the hybrid version of \#SAT, known as \#SMT \cite{approximate-counting-in-smt-and-value-estimation}. Since that proposal, numerous advances have been made on building efficient WMI solvers  (e.g., \cite{morettin2019advanced,merrell2017weighted,zeng2019efficient}) including the development of compilation targets \cite{ijcai18kolb,kolb2019pywmi,zuidberg2018knowledge}. 

% ,  albeit not as effective as propositional ones (yet) \cite{on-probabilistic-inference-by-weighted-model}.

Note that WMI proposes an extension of WMC for uncountably infinite (i.e., continuous) domains. What about countably infinite domains? The latter type is particularly useful for reasoning in (general) first-order settings, where we may say that a property such as \( \forall x,y,z ({\it parent}(x,y) \land {\it parent}(y,z) \supset {\it grandparent}(x,z)) \) applies to every possible \( x, y\) and $z$. Of course, in the absence of the finite domain assumption, reasoning in the first-order setting suffers from undecidability properties, and so various strategies have emerged for reasoning about an \textit{open universe} \cite{unifying-logic-and-probability}.  One popular approach is to perform \emph{forward reasoning}, where  samples needed for probability estimation are obtained from the facts and declarations in the probabilistic model  \cite{unifying-logic-and-probability,gutmann2011magic}. Each such sample corresponds to a possible world. But there may be (countably or uncountably) infinitely many worlds, and so exact inference is usually sacrificed. A second approach is to restrict the model wrt the query and evidence atoms and define estimation from the resulting finite sub-model  \cite{approximate-inference-for-infinite-contingent,markov-logic-in-infinite-domains,grohe2019probabilistic}, which may  also be substantiated with exact inference in special cases  \cite{open-universe-weighted-model-counting,weighted-model-counting-functions}. 

Given the successes of logic-based solvers for inference and probability estimation, one might wonder whether  such solvers would also be applicable to learning tasks in models with relational features and hard, deterministic constraints? 
 These, in addition to other topics, are considered  in the next section.

% section logic_vs_machine_learning (end)

\section{Machine Learning for Logic} % (fold)
\label{sec:machine_learning_for_logic}

At least since the time of Socrates, inductive reasoning has been a core issue for the logical worldview, as we need a mechanism for obtaining axiomatic knowledge. In that regard, the learning of logical and symbolic artifacts is an important issue in AI,  and computer science more generally \cite{dimensions-in-program-synthesis}.  There is a considerable body of work on learning propositional and relational formulas, and in context of probabilistic information, learning weighted formulas \cite{benedikt2020logic,inductive-logic-programming:-theory,de2015inducing,raedt2016statistical}. Approaches can be broadly lumped together as follows.  \begin{enumerate}
	\item \emph{Entailment-based   scoring:} Given a logical language \( \L, \) background knowledge \( \B \subset \L, \) examples \( \D \) (usually a set of \( \L \)-atoms), 
		find a hypothesis \( \H \in {\overline \H}, \H \subset \L \) such that \( \B \cup \H  \) entail the instances in \( \D. \)
	Here, the set \( {\overline \H} \) places  restrictions of the syntax of \( \H \) so as to control model complexity and generalization. (For example, \( \H = \D \) is a trivial hypothesis that satisfies the entailment stipulation.)
	
	\item \emph{Likelihood-based scoring:} Given \( \L,\B \) and \( \D \) as defined above, find \( \H\subset \L \) such that \( {\it score}(\H, \D) \gt {\it score}(\H', \D) \) for every \( \H' \neq \H. \) As discussed before, we might define \( {\it score}(\H ,\D) \propto \log \Pr(\D\mid \H) - {\it size}(\H) \). Here, like \( {\overline \H} \) above, \( {\it size}(\H) \) attempts to the control model complexity and generalization. 
	
\end{enumerate}

Many recipes based on these schemes are possible.   For example, we may use entailment-based inductive synthesis for an initial estimate of the hypothesis, and then resort to Bayesian scoring models \cite{markov-logic-networks}. The synthesis step might invoke neural machinery \cite{evans2018learning}. We might not require that the hypothesis entails every example in \( \D \) but only the largest consistent subset, which is sensible when  we expect the examples to be noisy  \cite{de2015inducing}. We might compile \( \B \) to an efficient data structure, and perform likelihood-based scoring on that structure \cite{liang2017learning}, and so \( \B \) could be seen as deterministic domain-specific constraints. Finally, we might stipulate the conditions under which a ``correct'' hypothesis may be inferred wrt unknown ground truth, only a subset of which is provided in \( \D. \) This is perhaps best represented by the  (probably approximately correct) PAC-semantics  that captures the quality possessed by the output of learning algorithm whilst costing for the number of examples that need to be observed \cite{valiant2000robust,cohen1994pac}. (But other formulations are also possible, e.g., \cite{grohe2017learning}.) 

This discussion pertained to finite domains.  
What about continuous spaces? By means of arithmetic fragments and 
formulations like WMI, it should be clear that  it now becomes possible to extend the above schemes to learn continuous properties. For example, one could learn linear expressions from data \cite{kolb2018learning}. For an account that also tries to evaluate a hypothesis that is correct wrt unknown ground truth, see \cite{mocanu2019pac+}. 
If the overall objective is to obtain a distribution of the data, other possibilities present themselves. 
In \cite{nitti2016learning}, for example, real-valued data points are first  lumped together to   obtain 
atomic continuous random variables.  From these,  relational formulas  are constructed so as to yield hybrid probabilistic programs. The learning is based on likelihood scoring. In  \cite{speichert2018learning}, the real-valued data points are first intervalized, and polynomials are learned for those intervals  based on  likelihood scoring. These weighted atoms are then used for  learning clauses by entailment judgements \cite{de2015inducing}. 

Such ideas can also be extended to data structures inspired by knowledge compilation, often referred to as \textit{circuits}  \cite{on-probabilistic-inference-by-weighted-model,sum-product-networks:-a-new-deep-architecture}. Knowledge compilation \cite{DBLP:journals/jair/DarwicheM02} arose as a way to represent logical theories in a manner where certain kinds of computations (e.g., checking satisfiability)  is significantly more effective, often polynomial in the size of the circuit. In the context of probabilistic inference, the idea was to then position probability estimation to  also be computable in time polynomial in the size of the circuit \cite{on-probabilistic-inference-by-weighted-model,sum-product-networks:-a-new-deep-architecture}. Consequently, (say) by means of likelihood-based scoring, the learning of circuits is particularly attractive because once learned, the bottleneck of inference is alleviated \cite{learning-arithmetic-circuits,liang2017learning}. In 
\cite{molina2018mixed,bueff2018tractable}, along the lines of the work above on learning logical formulas in continuous domains, it is shown that the learning of circuits can also be coupled with WMI.  

%, where certain kinds of computations (e.g., probability estimation) are significantly more effective .  

What about countably infinite domains? In most pragmatic instances of learning logical artifacts, the difference between the uncountable and countably infinite setting is this: in the former, we see finitely many real-valued samples as being drawn from an (unknown) interval, and we could  inspect these samples to crudely infer a lower and upper bound. In the latter, based on finitely many relational atoms, we would need to infer a universally quantified clause, such as \( \forall x,y,z ({\it parent}(x,y) \land {\it parent}(y,z) \supset {\it grandparent}(x,z)) \). If we are  after a hypothesis that is simply guaranteed to be consistent wrt the observed examples, then standard rule induction strategies would suffice \cite{inductive-logic-programming:-theory}, and we could interpret the rules as quantifying over a countably infinite domain. But this is somewhat unsatisfactory, as there is no distinction between the rules learned in the standard finite setting and its supposed applicability to the infinite setting. What is really needed is an analysis of what rule learning would mean wrt the infinitely many examples that have \emph{not} been observed. This was recently considered via the PAC-semantics in  \cite{belle2019implicitly}, by appealing to ideas on reasoning with open universes discussed earlier \cite{open-universe-weighted-model-counting}. 

% Although PAC-semantics is distinguished in being able to provide guarantees of generalization performance and polynomial time complexity, it should be noted that in the above result, no explicit hypothesis is actually produced, and the framework currne
%
% It is worth noting that there are significant differences between  PAC-semantics approaches and the others considered in this work, such as in terms of the learning regime, the notion of correctness, and the underlying algorithmic machinery

% Incidentally, the criticism we make of inducing hypothesis that is only consistent wrt the observed examples can be extended also to the continuous setting. There too, based on finitely many real-valued samples, we induce a  interval hypothesis. Can the PAC-semantics account be lifted to  continuous spaces as well. Results in \cite{mocanu2019pac+} answer in the affirmative. 

%Thus, by combining WMI and such structures, we obtain a tr

% If the linear expressions were given, one could learn the weights on expressions by density estimation strategies

Before concluding this section, it is worth noting that although the above discussion is primarily related to the learning of logical artifacts, it can equivalently be seen as a class of machine learning methods that leverage symbolic domain knowledge \cite{domingos2015master}. 
%machine learning can, of course, be influenced by domain knowledge. 
Indeed,  logic-based probabilistic inference  over deterministic constraints,  and  entailment-based induction augmented with  background knowledge are instances of such a class. 
%it already provides a  means for the underlying learning methodology to benefit from domain knowledge. 
Analogously, the automated construction of relational and  statistical knowledge bases \cite{deepdive:-web-scale-knowledge-base-construction,toward-an-architecture-for-never-ending-language}  by combining background knowledge with extracted tuples (obtained, for example, by applying natural language processing techniques to large textual data) is  another instance of such a class.  
%Nonetheless, the resulting artifact is a primarily a structure that is perhaps of most interest to a logician, or a knowledge representation scientist. 

In the next section, we will consider yet another way in which logical and symbolic artifacts can influence learning: we will see how such artifacts are useful to  enable tractability, correctness, modularity and compositionality.

% serve as meta-theory for machine learning to enable correctness, modularity and compositionality. 

%and the interfacing of reasoning and learning. 

%These and other points will be discussed in the next section. 

% See \cite{inductive-logic-programming:-theory,de2015inducing,raedt2016statistical} for discussions. For example, inspired by knowledge compilation approaches for probabilistic inference, the learning of  circuit languages is emerging as a promising area of research that is especially competitive in the presence of deterministic constraints \cite{liang2017learning}.

% section machine_learning_for_logic (end)

\section{Logic for Machine Learning} % (fold)
\label{sec:logic_for_machine_learning}

There are two obvious ways in which a logical framework can provide insights on machine learning theory. First, consider that computational tractability is of central concern when applying logic in computer science,  knowledge representation, database theory and search \cite{tractable-reasoning-with-incomplete,expressiveness-and-tractability-in-knowledge-representation,a-framework-for-representing-and-solving-np-search}.  Thus, the natural question to wonder is whether these ideas would carry over to probabilistic machine learning. On the one hand, probabilistic extensions to tractable knowledge representation frameworks could be considered \cite{p-classic:-a-tractable-probablistic-description}. But on the other, as discussed previously, ideas from knowledge compilation, and the use of circuits, in particular, are proving very effective for designing tractable paradigms for machine learning. While there has always been an interest in capturing tractable distributions by means of low tree-width models \cite{bach2002thin}, knowledge compilation has provided a way to also represent high tree-width models and enable exact inference for a range of queries \cite{sum-product-networks:-a-new-deep-architecture,liang2017learning}. See \cite{darwiche2020three} for a comprehensive view on the use of knowledge compilation for machine learning. 

%Perhaps the most obvious 
The other obvious way logic can provide insights on machine learning theory is by offering a formal apparatus to reason about \textit{context}. 
Machine learning problems are often positioned as atomic tasks, such as a classification task where regions of images need to be labeled as cats or dogs. However, even in that limited context, we imagine the resulting classification system as being deployed as part of a larger system, which includes various modules that communicate or interface with the classification system. We imagine an implicit accountability to the labelling task in that the detected object is either a cat or a dog, but not both. If there is information available that all the entities surrounding the object of interest have been labelled as lions, we would want to accord a high probability to the object being a  cat, possibly a wild cat. There is a very low chance of the object being a dog, then. If this is part of a vision system on a robot, we should ensure that the robot never tramples on the object, regardless of whether it is a type of cat or a dog. To inspect such patterns, and provide meta-theory for machine learning, it can be shown that symbolic, programmatic and logical artifacts are enormously useful. We will specifically consider correctness, modularity and compositionality to explore the claim.

% If a probabilistic machine learning model has to be defined wrt domain-specific constraints, perhaps 

%These have been investigated, and indeed are perhaps best motivated, in a finite setting, although  

%(These have been primarily investigated in a finite setting, although there 
%We note that what follow has been primarily investigated,  and indeed is perhaps best motivated, in a finite setting, although 

On the topic of correctness, the classical framework in computer science is \emph{verification}: can we provide a formal specification of what is desired, and 
can the system be checked against that  specification? In a machine learning context, we might  ask whether the system, during or after  training, satisfies a specification. 
% More generally, taking into account perturbations in the input space and training regime, can we ensure that the system robustly satisfies the specification?
The specification here might mean  constraints about the physical laws of the domain, or notions of perturbation in the input space while ensuring that the labels do not change, or  insisting that the prediction does  not label an object as being both a cat and a dog, or otherwise ensuring that outcomes   are not subject to, say, gender bias. Although there is a broad body of work on such issues, touching more generally on \textit{trust} \cite{rudin2018optimized}, we discuss approaches closer to the thrust of this article. For example, \cite{huang2017safety} show that a trained neural network can be verified by means of an SMT encoding of the network. In recent work, \cite{xu2018semantic} show that the loss function of deep learning systems can be adjusted to logical constraints by  insisting that the distribution on the predictions is proportional to the weighted model count of those constraints. In \cite{liang2017learning}, prior (logical) constraints are compiled to a circuit to be used for probability estimation. 
In \cite{papantonis2020constraint}, circuits are shown to be amenable to training  against  probabilistic and causal prior constraints, including assertions about fairness, for example. 

In \cite{solving-probability-problems-in-natural,manhaeve2018deepproblog}, a somewhat different approach to respecting domain constraints is taken: the low-level prediction is obtained as usual from a machine learning module, which is then interfaced with a probabilistic relational language and its symbolic engine. That is, the reasoning is positioned to be tackled directly by the symbolic engine. In a sense, such approaches cut across the three strands: the symbolic engine uses weighted model counting, the formulas in the language could be obtained by (say) entailment-based scoring, and the resulting language supports modularity and compositionality (discussed below).  

While there is not much to be said about the distinction between finite vs infinite wrt  correctness, many of these ideas    are likely amenable to extensions to an infinite setting in the ways  discussed in the previous sections (e.g., considering constraints of a  continuous or a countably infinite nature).

On the topic of modularity, recall that the general idea is  to reduce, simplify or otherwise abstract a (probabilistic) computation as an atomic entity, which is then to be referenced in another, possibly more complex, entity. In standard programming languages, this might mean the compartmentalization and interrelation of computational entities. For machine learning, approaches such as probabilistic programming \cite{church:-a-language-for-generative-models,de2015probabilistic} support probabilistic primitives in the language, with the intention of making  learning modules re-usable and modular. It can be shown, for example, that the computational semantics of some of these languages reduce to WMC    \cite{inference-in-probabilistic-logic-programs,holtzen2020dice}. Thus, in the infinite case, a corresponding reduction to WMI follows  \cite{dos2019exact,speichert2018learning,quantifying-program-bias}. 

A second dimension to modularity is the notion of \emph{abstraction}. Here, we seek  to model, reason and explain the behavior of systems in a more tractable search space, by omitting irrelevant details. The idea is widely used in natural and social sciences. Think of understanding the political dynamics of elections by studying micro level phenomena (say, voter grievances in counties) versus macro level events (e.g., television advertisements, gerrymandering). In particular, in computer science, it is often understood as the process of mapping one representation onto a simpler representation by suppressing irrelevant information. 
%Incidentally, when considering WMI, it can be noted that piecewise densities  are abstracted to a set of weighted propositions \cite{probabilistic-inference-in-hybrid-domains,dos2019exact}. 
% In the simplest instance, it is curious to note   piecewise densities  can be abstracted to a set of weighted propositions, as leveraged in WMI \cite{probabilistic-inference-in-hybrid-domains,dos2019exact}. 
%A more general observation is that abstraction allows us to process (categorical and probabilistic information) at the required granularity.
In fact, integrating low-level behavior with high-level reasoning, exploiting  relational representations to reduce the number of inference computations, and many other search space reduction techniques can all loosely be seen as instances of abstraction \cite{belle2020abstracting}. 

While there has been significant work on abstraction in deterministic systems \cite{banihashemi2017abstraction}, for machine learning, however, a probabilistic variant is clearly needed. In \cite{holtzen2017probabilistic}, an account of abstraction for loop-free propositional probabilistic programs is provided, where certain parts of the program (possibly involving continuous properties) can be reduced to a Bernoulli random variable. For example, suppose every occurrence of the continuous random variable $x$, drawn uniformly on the interval [0,1], in a program is either of the form $x\leq 7$ or of the form $x\gt 7$. Then, we could use a discrete random variable $b$ with a 0.7 probability of being true to capture $x\leq 7$; and analogously, $\neg b$ to capture $x\gt 7$. The resulting program is likely to be simpler. 
%Here too, the underlying computation can be seen as a WMC task \cite{holtzen2020dice}. 
In \cite{belle2020abstracting}, an account of abstraction for probabilistic relational models is considered, where the notion of abstraction also extends to deterministic constraints and complex formulas. For example, a single probabilistic variable in the abstracted model could denote a complex logical formula in the original model. Moreover, the logical properties that enable  verifying and inducing abstractions are also considered, and it is shown how WMC is sufficient for the computability of these properties (also see \cite{holtzen2020dice}). 

Incidentally, abstraction brings to light a reduction between finite vs infinite: it is shown in \cite{belle2020abstracting} that the modelling of piecewise densities as weighted propositions, which is leveraged in WMI \cite{probabilistic-inference-in-hybrid-domains,dos2019exact}, is a simple case of  the more general account. Therefore, it is worthwhile to investigate whether this or other accounts of abstraction  could emerge as  general-purpose tools that allow us to inspect the conditions under which infinitary  statements reduce to finite  computations. 

%(The abstraction of piecewise densities as weighted propositions, which is leveraged in WMI \cite{probabilistic-inference-in-hybrid-domains,dos2019exact}, turns out to be one simple case of  this more general account.) 

% It is shown that abstraction can be studied using notions such as WMC (also see \cite{holtzen2020dice}), and moreover, the logical properties that enable   verifying and inducing abstractions are further analyzed. The WMI reduction to WMC, for example, is one special case of this more general account. In that spirit, it is conceivable that abstraction might emerge as a general-purpose tool that allows us to inspect the conditions under which infinitary statements reduce to finite computations. 

% Interestingly, {although we consider logical artifacts for correctness, modularity and compositionality (below), 

A broader point here is the role abstraction might play in generating explanations \cite{explainable-artificial-intelligence-xai}. For example, a user's understanding of the domain is likely to be different from the low-level data that a machine learning system interfaces with~\cite{sreedharan2018hierarchical}, and so, abstractions can capture these two levels in a formal way.

Finally, we turn to the topic of compositionality, which, of course, is closely related to modularity in that we want to distinct modules to come together to form a complex composition. 
Not surprisingly, this is of great concern in AI, as it is widely acknowledged that most AI systems will involve heterogeneous components, some of which may involve learning from data, and others reasoning, search and symbol manipulation \cite{marcus2019rebooting}. In continuation with the above discussion, probabilistic programming is one such endeavor that purports to tackle this challenge by allowing modular components to be composed over programming and/or logical connectives \cite{church:-a-language-for-generative-models,de2015probabilistic,markov-logic-networks,manhaeve2018deepproblog,solving-probability-problems-in-natural,nitti2017planning,bundy,allegro:-belief-based-programming-in-stochastic,reasoning-about-uncertainty,logic-meets-probability:-towards}. (See  \cite{a-framework-for-representing-and-solving-np-search,modular-systems-with-preferences,an-abstract-view-on-modularity-in-knowledge} for ideas in deterministic systems.) However, probabilistic programming only composes probabilistic computations, but does not offer an obvious means to capture other types of search-based computations, such as SAT,  and integer and convex programming. 

Recall that the computational semantics of probabilistic programs reduces to WMC   \cite{inference-in-probabilistic-logic-programs,holtzen2020dice}. Following works such as \cite{semiring-based-constraint-logic-programming:,dyna:-extending-datalog-for-modern}, an interesting observation made in \cite{algebraic-model-counting} is that by appealing to a sum of products computation over different semiring structures, we can realize a large number of tasks such as satisfiability, unweighted model counting, sensitivity analysis, gradient computations, in addition to WMC.  
It was then shown in \cite{belle2016semiring} that the idea could be generalized further for infinite domains: by defining a measure on first-order models, WMI and convex optimization can also be captured. As the underlying language is a logical one, composition can already be defined using logical connectives. But an additional, more involved, notion of composition is also proposed, where a sum of products over different semirings can be concatenated. To reiterate, the general idea behind these proposals \cite{dyna:-extending-datalog-for-modern,algebraic-model-counting,belle2016semiring} 
is to arrive at a principled paradigm that allows us to interface learned modules with other types of search and optimization computations for the compositional building of AI systems. See  also \cite{kordjamshidi2018systems} for analogous discussions, but where a different type of coupling for the underlying computations is suggested. Overall, we observed that a formal apparatus (symbolic, programmatic and logical artifacts) help us define such compositional constructions by providing a meta-theory. 

%Overall, in all cases above, 

\section{Conclusions} % (fold)
\label{sec:conclusions}

In this article, we surveyed work that provides further evidence for the connections between logic and learning. 
Our narrative was structured in terms of three strands: logic versus learning, machine learning for logic, and logic for machine learning, but naturally, there was considerable overlap.

%Indeed, it should be clear that what we are really advocating for is the tackling of problems that symbolic logic and machine learning might struggle to realize individually
We covered a large body of work on what these connections look like, including, for example, pragmatic concerns such as the use of hard,  domain-specific constraints and background knowledge, all of which considerably eases the requirement that all of the agent's knowledge should be derived from observations alone. (See discussions in \cite{levesque2017common} on the limitations of learned behavior, for example.)  
Where applicable, we placed an emphasis on how extensions to infinite domains are possible. In the very least, logical artifacts can help in constraining, simplifying and/or composing machine learning entities, and in providing a principled way to study the underlying representational and computational issues. 

In general, this type of work could help us move beyond the narrow focus of the current learning literature so as to deal with time, space, abstraction, causality, quantified generalizations, relational abstractions, unknown domains, unforeseen examples, among other things, in a principled fashion. In fact, what is being advocated is the tackling of problems that symbolic logic and machine learning might struggle to address individually.  One could even think of the need for a recursive combination of strands 2 and 3: purely reactive components interact with purely cogitative elements, but then those reactive components are learned against domain constraints, and the cogitative elements are induced from data, and so on. More broadly, making progress towards a formal realization of \emph{System 1} versus \emph{System 2} processing might also contribute to our understanding of human intelligence, or at least capture human-like intelligence in automated systems. 

% More broadly, making progress towards a formal realization of \emph{System 1} versus \emph{System 2} processing would likely be of interest also to cognitive science and our understanding of human intelligence.

% section conclusions (end)

% (A broader point is the role abstraction might play in generating explanations, as a user's understanding of the domain is likely to be different from the low-level )

%It is also conceivable that abstraction might play a key role in generating explanations, as a user's understanding 

% WMC abstraction 

% \footnote{We note that in the sequel, we will not  draw out the distinction between finite vs infinite, as there are larger issues that are more significant. Nonetheless, many of the ideas discussed  below are amenable to extensions to the infinite setting in the ways already discussed.}

%  respect the presence of physical paths between the nodes on the map
%
%
%

% section logic_for_machine_learning (end)
%Although discrete, can be continuous. Next section.

%(These have been investigated, and indeed are perhaps best motivated, in a finite setting, although 

\bibliographystyle{abbrv}
\bibliography{group}
\end{document}